\documentclass[letterpaper, 10 pt, journal, twoside]{ieeetran}

\usepackage{graphics} 
\usepackage{color}
\usepackage{subcaption}
\usepackage[british]{babel}
\usepackage{multirow}
\usepackage[hyphens]{url}
\usepackage{hyperref}
\usepackage{cite}
\usepackage{epsfig} 
\usepackage{amsmath,amssymb}
\usepackage{multirow}
\usepackage{algorithm,algorithmic}
\usepackage{subcaption}
\usepackage{multirow}
\usepackage{url}
\usepackage{comment}
\usepackage[table]{xcolor}

\DeclareMathOperator*{\argmax}{arg\,max}

\graphicspath{ {./images/} }
\DeclareGraphicsExtensions{.png,.pdf,.jpeg,.jpg}

\newcommand{\maxv}{n}

\newcommand{\tset}{T}
\newcommand{\vset}{V}

\newcommand{\ainA}[2]{#1 \in #2} 				

\newcommand{\tost}[2]{{#1}^{a, t#2}} 				
\newcommand{\bcap}{b_{\rm c}(t)}

\newcommand{\figc}[1]{Fig. \ref{#1}}

\newcommand{\dangerhat}{{\boldsymbol{\eta}}^{t}_{v}}

\newcommand{\dangerpt}{{z}^{t}_{v}}
\newcommand{\dangerprob}{{{H}}^{a, t}_{v}}

\newcommand{\dthres}{\kappa^{a}}
\newcommand{\dconf}{\alpha^{a}}

\newcommand{\ploss}{p(\text{loss}|l_v)}

\newcommand{\moderate}{moderate}
\newcommand{\substantial}{high}
\newcommand{\severe}{very high}
\newcommand{\critical}{extreme}

\newcommand{\level}{l}
\newcommand{\lset}{\mathcal{L}}

\newcommand{\mywc}{2.7cm}
\newcommand{\myha}{2.45cm}

\newcommand{\makeup}{makeup}

\usepackage{todonotes}
\usepackage{soul}
\definecolor{smoothgreen}{rgb}{0.7,1,0.7}
\sethlcolor{smoothgreen}

\newcommand{\todohere}[1]{\hl{(\textbf{TODO:} #1)}}

\definecolor{darkpink}{rgb}{0.91, 0.33, 0.5}
\newcommand {\vikram}[1]{{\color{black}#1}}
\newcommand {\bea}[1]{{\color{black}#1}}
\newcommand {\brev}[1]{{\color{black}#1}}

\newcommand {\jacopo}[1]{{\color{black}#1}}

\newcommand {\revb}[1]{{\color{black}#1}}

\title{Exploiting Natural Language for Efficient Risk-Aware Multi-robot SaR Planning}

\author{Vikram Shree$^{*}$, Beatriz Asfora$^{*}$, Rachel Zheng, Samantha Hong, Jacopo Banfi, and Mark Campbell
\thanks{*Equal contribution, order decided randomly. All authors are with the Sibley School of Mechanical and Aerospace Engineering, Cornell University, Ithaca, NY USA. Email: {\tt\small \{vs476, ba386, rz246, sh974, jb2639, mc288\}@cornell.edu}. 
}
\thanks{ Research supported by the NRI program of the National Science Foundation, award \#1830497.}

\thanks{\textbf{Cite as:} V. Shree, B. Asfora, R. Zheng, S. Hong, J. Banfi and M. Campbell, ``Exploiting Natural Language for Efficient Risk-Aware Multi-Robot SaR Planning,'' in IEEE Robotics and Automation Letters, vol. 6, no. 2, pp. 3152-3159, April 2021.}
\thanks{The official IEEE published version of this manuscript can be accessed at: \href{https://ieeexplore.ieee.org/abstract/document/9366368}{https://ieeexplore.ieee.org/abstract/document/9366368}
}
\thanks{\copyright 2021 IEEE. Personal use of this material is permitted. Permission from IEEE must be obtained for all other uses, in any current or future media, including reprinting/republishing this material for advertising or promotional purposes, creating new collective works, for resale or redistribution to servers or lists, or reuse of any copyrighted component of this work in other works.}
}

\begin{document}

\maketitle


\begin{abstract}
\label{sec:abstract}
The ability to develop a high-level understanding of a scene, such as perceiving danger levels, can prove valuable in planning multi-robot search and rescue (SaR) missions. In this work, we propose to uniquely leverage natural language descriptions from the  \bea{mission commander in chief} and image data captured by robots to estimate scene danger. Given a description and an image, a state-of-the-art \jacopo{deep neural network} is used to assess a corresponding similarity score, which is then converted into a probabilistic distribution of danger levels. Because commonly used visio-linguistic datasets do not represent SaR missions well, we collect a large-scale image-description dataset from synthetic images taken from realistic disaster scenes \jacopo{and use it to train our machine learning model}. A risk-aware variant of the Multi-robot Efficient Search Path Planning (MESPP) problem is then formulated \bea{to use the danger estimates in order to account for high-risk locations in the environment when planning the searchers' paths.} The problem is solved via a distributed approach based on Mixed-Integer Linear Programming.  
Our \revb{experiments} demonstrate that our framework allows to plan safer \bea{yet highly successful search missions, abiding to} the two most important aspects of SaR missions: to ensure both searchers' and victim safety. 
\end{abstract}
\section{Introduction}
\label{sec:introduction}

\IEEEPARstart{A}{ccurate} scene awareness is the keystone for success in search and rescue (SaR) missions. The deployment of robots in the  World Trade Center disaster pinpointed limitations in different aspects of robotic systems, including human robot collaboration \cite{murphy2003}. For the past two decades, sensors in particular have evolved in number and variety, and are now able  to generate gigabytes of data within seconds, and even extract important features autonomously. However, rigorous analysis and summarizing of the large amount of data about a scene in order to infer high-level understanding of the surrounding world is still a work in progress. Errors propagate across and down the multirobot system pipeline, with detrimental impact to the SaR mission performance. 

Prior work in robotic perception has focused on inferring low-level information about the environment, for example, building occupancy grid maps \cite{thrun2003learning} or mapping unique landmarks \cite{se2002mobile}. These low-level attributes are indeed relevant to build a representation of the world that is suitable for navigation. However, planning for a team of agents typically requires humans to make decisions based on high-level attributes of the environment. These include the notion of ``danger'', for which the ability to map low-level aspects of the scene into a high-level and succinct representation is still an open question in the robotics community. In this work, we address this problem by proposing a systematic approach for inferring scene danger from visio-linguistic inputs to enable high-level planning for a team of agents.

\begin{figure}
    \centering
    \includegraphics[trim= 0 0 0 0, clip, width=0.4\textwidth]{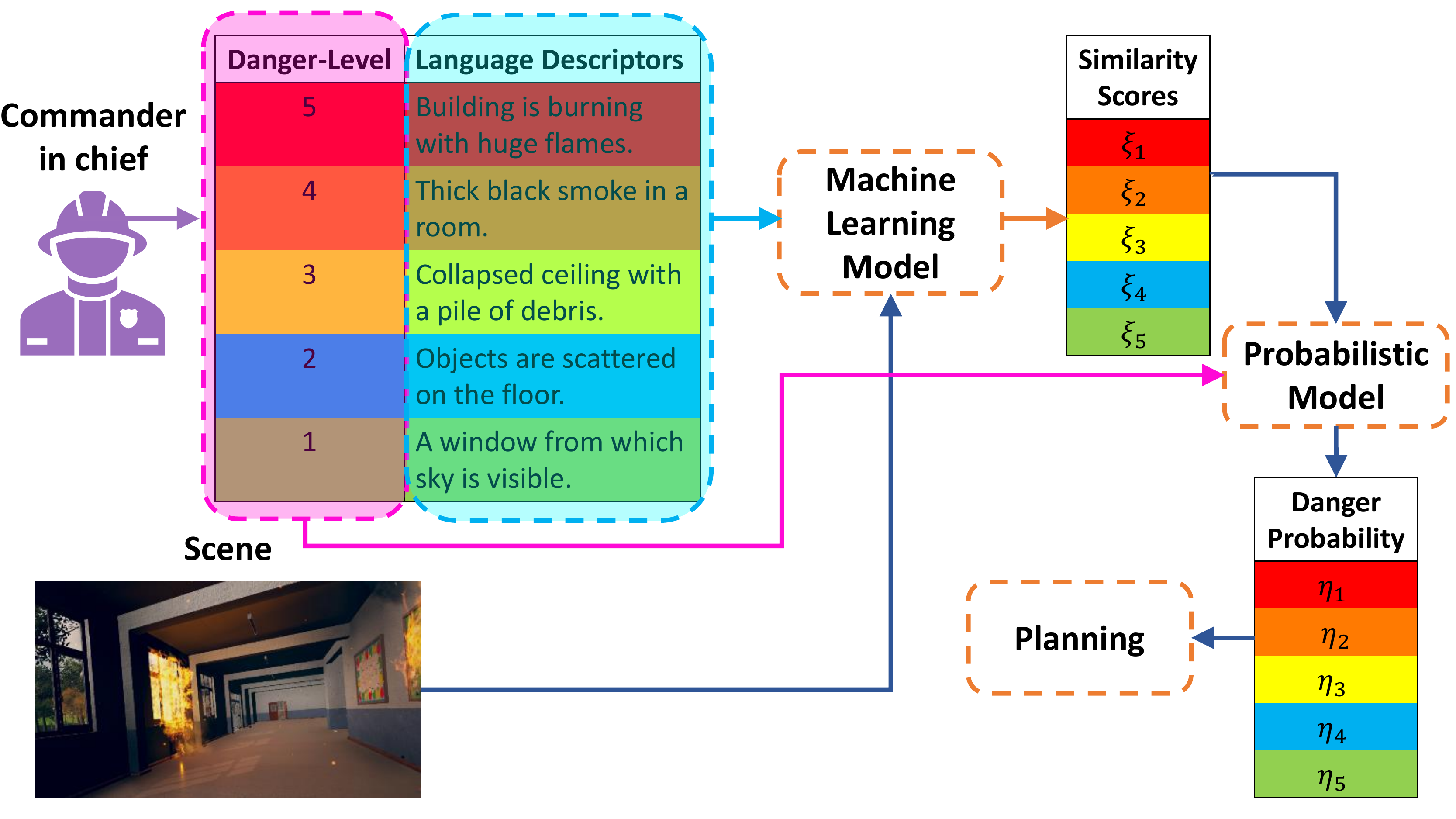}
    \caption{\small Scene danger perception and planning pipeline. A set of descriptors and corresponding danger levels, provided by the mission commander, is used for estimating danger probability distribution, on a 5-point scale. This information is used by the planning module to plan safe paths for the agents.}
    \label{fig:dangerEstimatePipeline}
\end{figure}


\jacopo{Our approach, sketched in Fig.~\ref{fig:dangerEstimatePipeline}, asks users for descriptions that they feel are important to characterize scene danger. Descriptions are matched against images seen from the robot's camera by leveraging a machine learning model, and the obtained similarity scores are used to keep up-to-date a probability distribution describing danger in different areas of the environment. This information, in turn, is used to plan more informed paths for the searchers. In summary, this paper makes the following novel contributions:}

\vikram{
\begin{enumerate}
    \item The adaptation of a state-of-the-art deep neural network \cite{lu2019vilbert} to assess similarity between the descriptions and images in the scene. To facilitate the use of the network in SaR missions, we introduce a novel dataset, consisting of language descriptions for synthetic disaster scenes taken from the DISC dataset \cite{jeon2019disc}.
    
    \item An intuitive probabilistic model for estimating scene danger by fusing similarity scores obtained from various user descriptions and multiple images at a scene. Compared against a system (e.g. a classifier) that is directly trained on a priori notions of ``danger'', our approach can adapt to the needs of different missions (e.g. fire, earthquake, or radioactive incident) without having to change any of its parameters. 
    
    \item The introduction of a risk-aware version of the Multi-robot Efficient Search Path Planning (MESPP) problem~\cite{hollinger2009} which can leverage scene understanding to ensure agents' safety by accounting for each heterogeneous agent's danger tolerance. This is an online problem, which we solve via a distributed planning approach based on variants of state-of-the-art Mixed-Integer Linear Programming (MILP) models \cite{asfora2020}.
    
\end{enumerate}
}

\jacopo{Extensive numerical and realistic ROS/Gazebo simulations show that our holistic approach to multi-robot SaR planning enforces the safety of heterogeneous agents under different notions of risks specified by the user, while maintaining performance in terms of time required to locate the victim.}

\section{Related Work}
\label{sec:relatedWork}

\vikram{\subsection{Collaborative search and rescue missions}

The challenges encountered in SaR missions depend upon the operational environment, which can be divided into three main categories: maritime, urban, and wilderness~\cite{queralta2020collaborative}. In this paper, we focus on SaR in urban environments. Robots can go into more dangerous areas to avoid risking human lives, however they still need to leverage the human expertise for scene understanding. Yazdani et al. \cite{yazdani2017cognition} studied how high-level instructions from a human can be used to plan actions for the robot during SaR missions. In order to bridge the gap between scene perception and decision-making, researchers have proposed ontology models \cite{sun2019high} that consist of rules to represent the environment and action space. However, these approaches have been only shown to work in simple settings and are prone to failure in more complex, uncertain scenarios, which are often encountered in SaR missions.
}

\subsection{Language-based scene assessment}
In a typical rescue mission, human rescuers talk with each other on a low-bandwidth radio channel since it has virtually unlimited range. \brev{The} human brain is great at interrelating data from visual and language domains, but when it comes to neural networks, this task becomes more challenging due to the lack of a one-to-one relationship between the two inputs. 

Modern deep neural architectures \cite{lee2018stacked, lu2019vilbert} tend to address this challenge by extracting high-level features from each one of the inputs before jointly reasoning about them. The benefits of these models, however, have only been tested on datasets describing serene environments (e.g.~\cite{plummer2015flickr30k}) because of the constraints associated with replicating a realistic SaR scenario. In this work, we leverage photo-realistic synthetic images from disaster scenes~\cite{jeon2019disc} to conduct a large-scale survey. This allows us to obtain language data, which is vital for training state-of-the-art visual-language models before applying them in SaR missions.

Compared to other approaches for processing visio-lingual data that can be found in the computer vision community, \jacopo{like visual question answering \cite{antol2015vqa} and visual commonsense reasoning} \cite{zellers2019recognition}, caption-based image retrieval \cite{young2014image} is the most relevant from the standpoint of this work. This refers to the task of identifying an image from a large pool given a caption describing its content, thus, requiring to estimate similarity between images and text data.  
There is a rich line of work towards mapping images and sentences to a common semantic space for assessing image-text similarity. In this paper, we compare the usage of SCAN~\cite{lee2018stacked} and ViLBERT \cite{lu2019vilbert} on a new dataset built on top of the disaster scenes contained in the DISC dataset \cite{jeon2019disc}, \brev{due to} their superior performance reported in the literature \cite{lu2019vilbert}.

\subsection{Multirobot search}
We assess the advantages brought by our danger estimation pipeline by formulating and \bea{evaluating} a risk-aware version of a famous robotic search problem, the Multi-robot Efficient Search Path Planning (MESPP) problem introduced by Hollinger et al. \cite{hollinger2009}. In the original version of the MESPP problem, a team of robots is deployed in a graph-represented environment with the aim of locating a moving non-adversarial target within a given deadline. In this new \brev{online} version, graph vertices are associated \bea{with} probabilistic danger estimates, and robots are heterogeneous in terms of their tolerance to \brev{hazardous conditions}. \jacopo{This is different from the settings that can be found in the literature, which either focus on static threats \cite{yehoshua2016b}, or consider dynamic threats but in presence of a single agent that has to reach a given goal location while maximizing its chances of survival (without optimizing a second performance metric)~\cite{banfi2020planning}.}

\begin{figure*}[ht]
    \centering
    \vspace{0.1cm}
    \includegraphics[width=15.8cm, height=5cm]{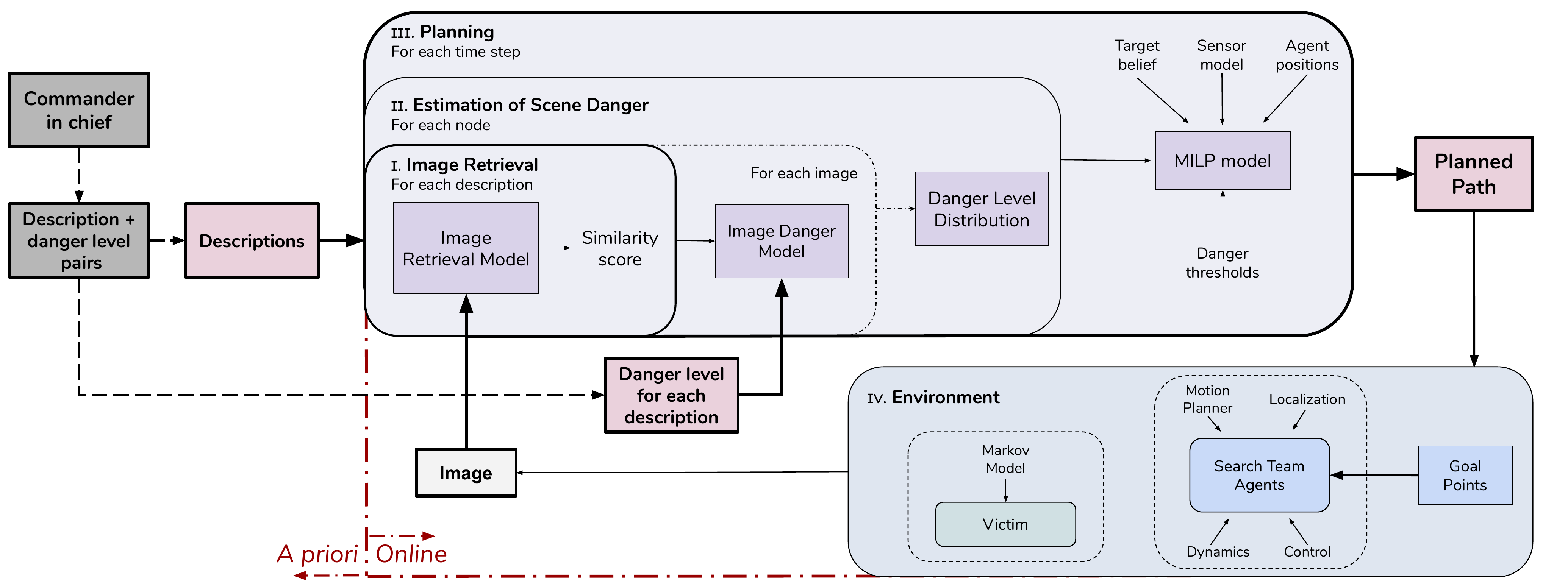}
    \caption{\brev{System layers: text-image similarity assessment (I), scene danger estimation (II) and multi-robot planning (III).}} 
    \label{fig:scheme}
    \vspace{-0.5cm}
\end{figure*}
\section{Proposed Architecture}
\label{sec:proposedArchitecture}

Let us define five intuitive danger levels
: {low}, {\moderate}, {\substantial}, {\severe}, and {\critical}, and assign a value $l \in \lset$ to each, respectively $\lset=\{1, ...,5\}$. We assume that an experienced user (the commander in chief) is able to provide at least one sentence characterizing a description of each danger level (e.g., ``a room is filled with fire'', associated with $l=5$). Our pipeline consists of three layers, as shown in Fig.~\ref{fig:scheme}. Layer I is tasked with estimating similarity between the user's descriptions and images collected in a scene, providing a matching score for each image-description pair. The scores of all images are combined into a probabilistic estimate of the danger level (Layer II), which is in turn used for planning an efficient risk-aware search for the team (Layer III). This plan is then sent to the team of robots operating in the environment, where they search for victim and acquire new images. Each layer is described in detail in the following sections.

\section{Text-Image Similarity Assessment}
\label{sec:simAssess}

As a first step to establish the user-perceived danger level of the scene, we start with determining similarity between the provided language descriptors and the scene. In this work, a scene consists of a set of images acquired online by the robot. The robot can compute a {\em similarity score} $\xi$ for each (image, descriptor) pair. This can be done with one of the following deep learning architectures.



\subsubsection{SCAN}
Stacked Cross Attention for Image-Text Matching (SCAN) \cite{lee2018stacked} uses a two-stage attention architecture for calculating text-image similarity. First, the word and image are converted into a set of word and image features, respectively. This is followed by calculating a cosine similarity matrix for each possible word-image feature pair. The first attention stage attends to image regions w.r.t each word in the sentence. The next stage compares the words based on the attended image vector and decides the importance of each word. The final pooling layer outputs a similarity score.

\subsubsection{ViLBERT}
Visual-Language BERT (ViLBERT) \cite{lu2019vilbert} introduces two separate streams for processing vision and language data that can communicate with each other via co-attention transformer layers. After having converted word and image into features, ViLBERT processes the features with transformer blocks independently before they could interact with each other. 
The co-attention layers produce attention features for each modality conditioned on the other. The final pooling layer outputs a similarity score.

\subsection{A Novel Emergency Scene Dataset: DISC-L}
\label{subsec:dataCollection}


The DISC-\vikram{Language} dataset\footnote{Freely available for academic purposes at \\ \href{https://github.com/vikshree/DISC-L.git}{https://github.com/vikshree/DISC-L.git}} is built on top of DISC datset~\cite{jeon2019disc}, consisting of 300K photo-realistic synthetic images taken in several environments (like office and subway) with two types of damage conditions: collapse and fire. The images are captured from a stereo-camera, moving along a pre-defined path in a 3D model world. To reduce redundancy in images, we uniformly sample 1 out of every 150 images from the left camera, yielding a set of 1000 images.

\begin{figure}
    \centering
    \includegraphics[trim= 0 0 0 160, clip, width=0.48\textwidth, height=3.7cm]{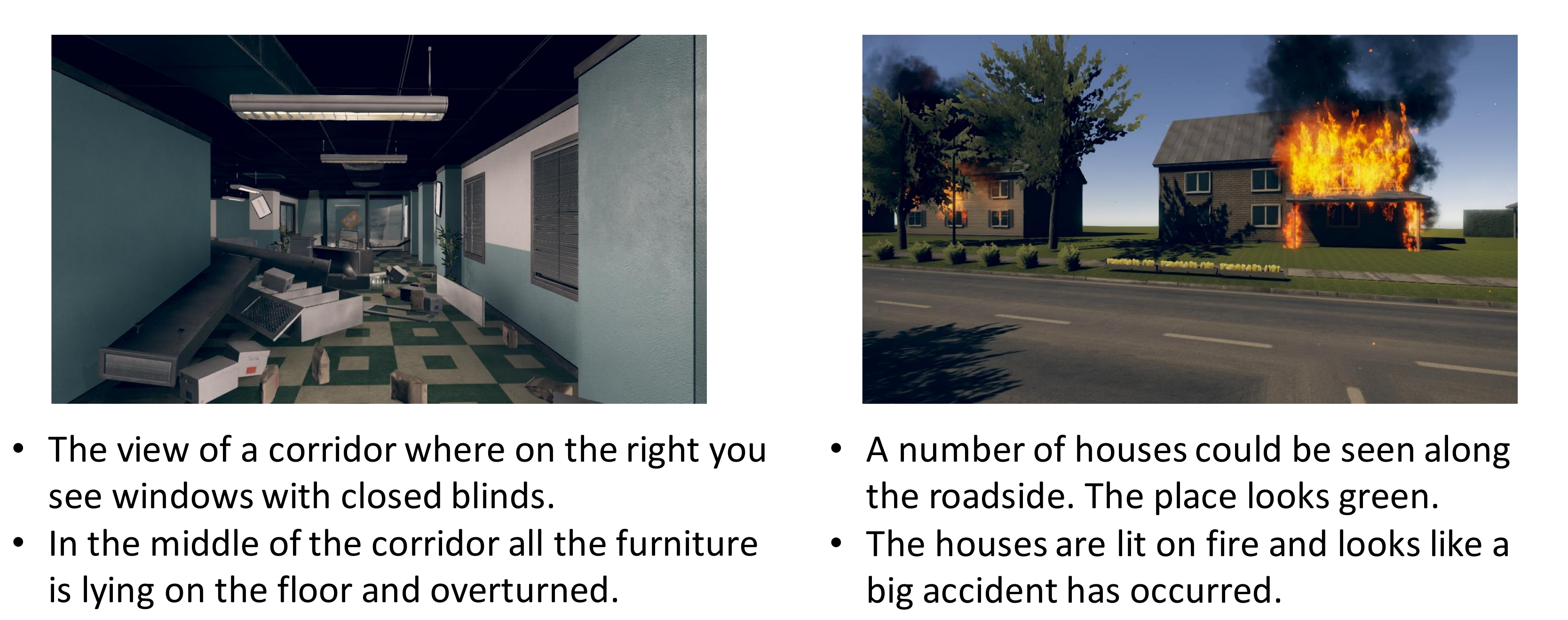}
    \caption{\small A few sample descriptions for images in the DISC-L dataset, collected through AMT survey.}
    \label{fig:discExamples}
    \vspace{-0.1in}
\end{figure}

We conduct an online survey on Amazon Mechanical Turk (AMT) to collect sentence descriptions for each image. \vikram{AMT is a global service enabling us to reach a huge participant pool which favorably introduces diversity in the collected data}. Each task requires the user to read the instructions, look at annotated examples, and describe important aspects of a given image with two sentences in English. To incorporate diversity in language, each image was labelled by four unique workers, yielding a total of 4000 descriptions. \vikram{Furthermore, to ensure high quality language input from the users, the responses are first filtered automatically and then manually approved to remove unsatisfactory descriptions.} A few examples are shown in Fig.~\ref{fig:discExamples}. \vikram{More examples of accepted and rejected responses can be found on the DISC-L Github page.} 


In summary, DISC-L dataset consists of rich and diverse descriptions of emergency situations from 720 unique Amazon users. There are a total of 3,386 unique words in our proposed dataset. \vikram{The word-count ranges from 10 words to 70 words per description, with a median of 20. A key observation is that the most frequently used words (fire, smoke, flame, dark etc.) are related to situations commonly encountered during SaR mission.}


\subsection{Caption-based Image Retrieval Performance}
\label{subsec:imageRetieval}

\subsubsection{Dataset} The DISC-L dataset is used to evaluate the performance of SCAN and ViLBERT. The dataset is divided into train, validation and test sets such that all images corresponding to a environment remain in the same set. The train set consists of 828 images from six environments; validation has 75 images from three environments; and test has 97 images from two environments. 

\subsubsection{Training} SCAN is pre-trained for caption-based image retrieval on the Flickr30K dataset \cite{plummer2015flickr30k}. \jacopo{A bottom-up attention network \cite{anderson2018bottom} 
is used to compute} a $36 \times 2048$ dimensional feature vector for each image, before feeding it to the SCAN network. 
Finally, the data batch is shuffled online to get negative sentence-image pairs, and a triplet loss function is used to fine-tune the network on the DISC-L dataset. \vikram{Triplet loss is a ranking-based loss function, commonly used in image-text matching task where distance between an anchor is minimized from the truth value (see~\cite{lee2018stacked} for details).}

ViLBERT is pre-trained for multiple visual-language tasks on 12 different datasets \cite{lu202012}.
A combination of pre-trained Faster R-CNN and ResNet \cite{he2016deep} is used to compute a $101 \times 2048$ dimensional feature vector for each image, which is fed to the ViLBERT network. Following the authors' approach \cite{lu2019vilbert}, we adopt a 4-way multiple choice training process where for each sentence-image pair in the dataset, three distractor pairs are sampled with no correspondences. 
Finally, we fine-tune the model on DISC-L dataset with a cross-entropy loss.

\subsubsection{Results} We use standard rank-based performance metrics (R-1, R-5, R-10) to evaluate the models, 
where R-$k$ denotes the proportion of times the correct image is present in the top $k$ likely hypotheses for a description. 

Table~\ref{tbl:retrievalPerformance} shows the results. Fine-tuning the models on DISC-L significantly improves the retrieval performance for both SCAN and ViLBERT, since the pre-trained models have never encountered danger-related images or sentences. 
Also, we observe that ViLBERT outperforms SCAN by a margin of $150\%$, $100\%$ and $79\%$ for R-1, R5 and R-10 respectively. The superior performance of ViLBERT can be attributed to its more complex model architecture and the use of higher dimensional feature representation for the images. Therefore, we use the ViLBERT model in the remainder of the paper.


\begin{table}
\vspace{0.1in}
\centering
\caption{ \small SCAN vs ViLBERT models evaluated on DISC-L.}
\begin{tabular}{|p{1.2cm}|p{2.7cm}|p{0.8cm}|p{0.8cm}|p{0.8cm}|}
\hline
  \textbf{Model} & \textbf{Training type} & \textbf{R-1} & \textbf{R-5}  & \textbf{R-10}\\ 
  \hline
 \multirow{2}{*}{\textbf{SCAN}} & Pre-trained &  7.5 & 24.0 & 39.2\\
 & Fine-tuned on DISC-L & 7.7 & 30.4 & 48.2\\
   \hline
 \multirow{2}{*}{\textbf{ViLBERT}} & Pre-trained &  9.7 & 39.3 & 59.7\\
 & Fine-tuned on DISC-L & \textbf{19.4} & \textbf{52.0} & \textbf{70.4}\\
 \hline
\end{tabular}
\label{tbl:retrievalPerformance}
\vspace{-0.1in}
\end{table}



\section{Scene Danger Estimation}
\label{sec:dangerCalculation}


\vikram{We formulate} the task of estimating a probability mass function over the danger space $\lset= \{1,...,5\}$, 
given a set of language descriptors from a user and a scene as a probabilistic inference task. A scene consists of a set of $r$ local images taken from a given region of the environment: $I = \{ i_1 , i_2, \ldots, i_r\}$. 

For \vikram{notational simplicity}, let us assume that a scene consists of a single image. A single set of descriptions is not sufficient to assess the danger level in different hazard conditions. Hence, our model allows the commander in chief to specify multiple descriptions for each level, which can then be grouped together based on $m$ danger ``types''. For example, those related to descriptions of a fire. In symbols, $S=\{S_1, S_2, \ldots, S_m\}$, where each set $S_j=\{s_j^1,\ldots,s_j^5\}$ contains a language descriptor for each of the 5 danger levels related to the $j$-th danger type. Given a single image $i$ and the full set of language descriptors $S$, the ViLBERT model is used to generate a matrix of similarity scores, denoted by $\Xi = \left\{ \{\xi_{1}^{1}, \ldots, \xi_{1}^{5}\}, \ldots,  \{\xi_{m}^{1}, \ldots,  \xi_{m}^{5}\} \right\}$. To reduce the impact of  noise, the scores are converted into one-hot vectors $\textbf{y} = \left\{ \textbf{y}_{1}, \ldots, \textbf{y}_{m} \right\}$ based on a threshold $\theta$, such that:
\begin{equation}
\begin{split}
    \textbf{y}_{u}^{l} = 1 \mbox{ if } \xi_{u}^{l} \geq \theta \mbox{ else } 0 \ \forall \ l \in \lset, \ u \in \{1,\ldots,m\}.
\end{split}
\label{eq:threshold}
\end{equation}
The posterior distribution of danger level for the scene $P(D|  \Xi, S, I)$, denoted by the vector $\boldsymbol{\eta} = [\eta_1 \; \eta_2 \cdots \eta_5]$, is calculated based on the frequency of samples in the data as:
\begin{equation}
    \eta_{l} = \frac{1}{\rho} \sum_{u=1}^{m} \textbf{y}_{u}^{l} \ \forall l \in \lset,
\label{eq:frequentist}
\end{equation}
where $\rho = \sum_{u=0}^{m} \sum_{l \in L} \textbf{y}_{u}^{l}$ is the normalizing constant. For the more general case where a scene consists of $r$ images, we simply include \vikram{one-hot vectors obtained} from each one of the images while calculating $\eta_{l}$ in Eq.~(\ref{eq:frequentist}).

\begin{figure}
    \centering
    \vspace{0.2cm}
    \includegraphics[width=0.491\columnwidth]{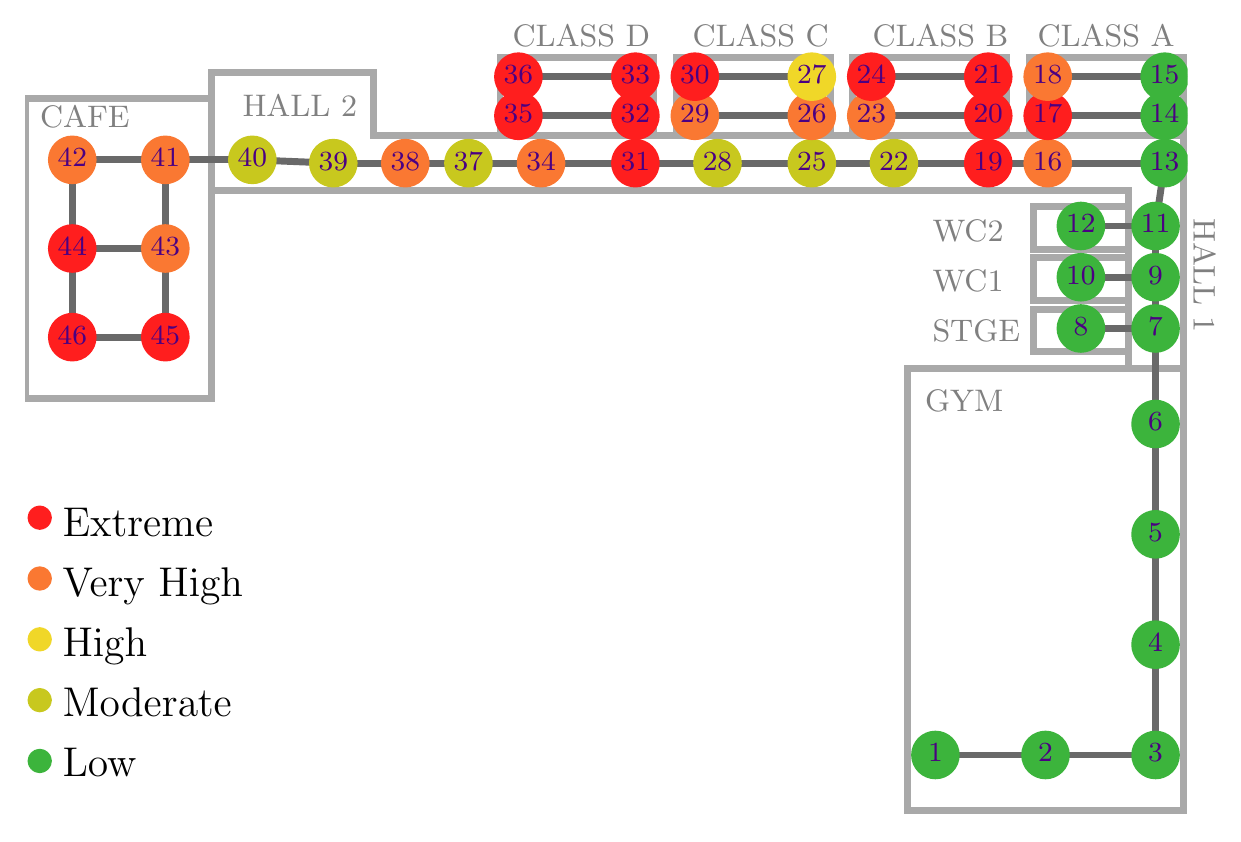}
    \includegraphics[width=0.491\columnwidth]{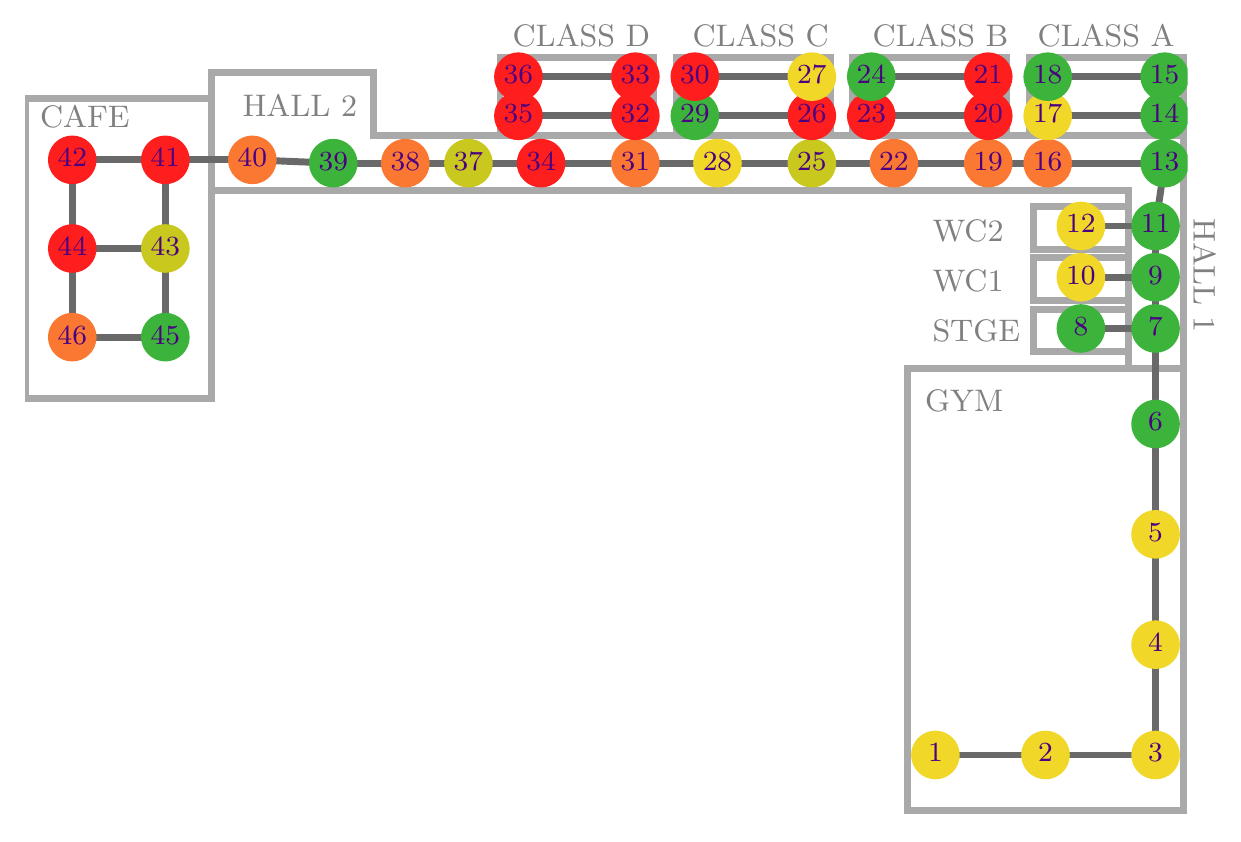}
    \caption{\brev{Graph abstraction of the school environment. Each vertex $v = 1,\ldots,46$ is color-coded based on its danger level in NFF scenario. Left: Human reasoned ground truth; Right: Estimate using 5\% of images and fire descriptors.}} 
    \label{fig:graph}
    \vspace{-0.1in}
\end{figure}

To evaluate our danger estimation approach, we use the School environment from the DISC dataset \cite{jeon2019disc} because of its relatively larger size and create the graph structure shown in Fig.~\ref{fig:graph}, containing $n=46$ vertices. Each vertex in the graph is associated with a set of local images collected from its immediate surroundings. The DISC dataset allows each vertex to be defined with images containing three different types of danger: none (N), collapse (C), and fire (F). Three `environments' are designed for simulation based on a predefined proportion of hazards: `NCF' denotes  environments where vertices are associated with  hazardous images in equal proportion; `NFF' denotes environments with no C-type danger and 2/3 of vertices with F-type danger, etc. We introduce a set of descriptions for assessing collapse and fire danger \jacopo{(see the DISC-L dataset Github page for details).} For example, the description ``The room is engulfed in huge flames'' describes a fire danger level $l=5$. Ground truth danger values $l_{v}$ for each vertex $v$ are obtained by having a human user reason about danger based on the descriptions and the images associated with vertex $v$.

We use the average Bhattacharya coefficient (BC)\vikram{\cite{comaniciu2000real}, a standard metric to quantify the disparity between two discrete probability distributions}, to \vikram{ measure the closeness} of the estimates with the ground truth, computed as:

\vspace{-2mm}
\begin{equation}
\begin{split}
    BC = \frac{1}{n} \sum_{v=1}^{n} \sqrt{\eta_{l_v, v}},
\end{split}
\label{eq:avg_bc}
\end{equation}

where $\eta_{l_v, v}$ is the probability corresponding to the ground-truth danger level $l_{v}$ at vertex $v$. \vikram{The results are shown in Table~\ref{tbl:dangerEstimation}.} First, we observe that the danger estimates for all environments are most accurate when the descriptions corresponding to that specific environment are used. For example, if it is a fire \vikram{hazard} (NFF), using only fire-related descriptions yields the best danger estimates ($BC=0.63$). If there are both fire and collapse vertices (NCF), then using a combination of both descriptors yields the best danger estimates ($BC=0.53$). Furthermore, the average BC obtained from the `\textit{uniform-prior}' baseline is 0.45, which is lower than the best estimates for each one of the scene-types: NFF (0.63), NCC (0.49), and NCF (0.53). Thus, we conclude that customizing descriptions\vikram{, depending on the hazard,} is indeed beneficial for danger estimation. \vikram{It should be noted that one can leverage homogeneous domain adaptation techniques~\cite{wang2018deep} to minimize any performance deterioration when transferring the system to the real world.} 





\begin{table}[H]
\vspace{-0.1in}
\centering
\caption{ \small Comparison of average BC for estimated danger distribution with different descriptions, for three scene types.}
\begin{tabular}{|p{1.4cm}|p{1.8cm}|p{1.8cm}|p{1.9cm}|}
\hline
  \textbf{Scene Type} & \textbf{F-Descriptors} & \textbf{C-Descriptors} & \textbf{FC-Descriptors}\\ 
  \hline
  \textbf{NFF} & \cellcolor{lightgray}\textbf{0.63} & 0.42 & 0.59\\ 
  \textbf{NCC} & 0.30 & \cellcolor{lightgray}\textbf{0.49} & 0.47\\ 
  \textbf{NCF} & 0.47 & 0.49 & \cellcolor{lightgray}\textbf{0.53}\\ 
 \hline
\end{tabular}
\label{tbl:dangerEstimation}
\end{table}

\section{Multi-robot Planning}
\label{sec:planning}

\subsection{Risk-Aware MESPP Problem}

\bea{We now introduce} a risk-aware version of the Multi-Robot Search Path Planning (MESPP) problem~\cite{hollinger2009}. In the original \bea{MESPP} problem, a team of robotic agents $A = \{1, ..., m\}$ is deployed in a graph-represented environment $G=(V,E)$, with the aim of locating a possibly moving non-adversarial ``target'' (e.g. a victim) within a given deadline $\tau$. Time $T = \{1, .., \tau\}$ evolves in discrete steps, and both the agents and the target can either stay at the same vertex or reach a neighbor vertex between two subsequent steps, for vertices $V = \{1,...,n\}$. Agents can communicate with each other at all time\bea{-steps}.

\bea{The target's probable motion in the graph is encoded by a Markovian matrix $\textbf{M} \in [0, 1]^{n\times n}$.} At each step $t$, the target's state, resulting from its interactions with agents executing a set of joint paths $\boldsymbol{\pi} \in \mathcal{P}$, is represented by the belief vector
\begin{equation}
\textbf{b}^{\boldsymbol{\pi}}(t) = [\bcap,~b_1(t),...,b_{\maxv}(t)],
\label{eq:beliefvector}
\end{equation}
\noindent where $\bcap + \sum_{i=1}^{n}b_{v}(t) = 1$. The first element, $\bcap$, represents the probability that the agents have located the target by time $t$ when following paths $\boldsymbol{\pi}$. The remaining elements $b_1(t),...,b_{\maxv}(t)$ represent the probability that the target is located in the corresponding vertices at time $t$. 

Detection events are described by matrices $\textbf{C}^{a,u} \in [0,1]^{(\maxv+1) \times (\maxv+1)}$, $\forall a \in A$, $\ainA{u}{\vset}$. Their effect is to connect the probability of the target being at a particular location with its detection state. In other words, the matrix $\textbf{C}^{a, u}$ encodes which vertices of the graph fall within the sensing range of agent $a$, when such is located in vertex $u$. A belief update equation links the current belief, the target's motion, and the agents' paths $\boldsymbol{\pi}$ with associated detection events as follows:
\begin{equation}
\textbf{b}^{\boldsymbol{\pi}}(t + 1)  = \textbf{b}^{\boldsymbol{\pi}}(t) \left[\begin{array}{cc}
1 & \textbf{0}_{1 \times n} \\ \textbf{0}_{n \times 1} & \textbf{M}\end{array} \right] \prod_{a \in A} {\textbf{C}}^{a, \pi^{a, t+1}}.
\label{eq:beliefupdate}
\end{equation}

\jacopo{In the original MESPP problem, the goal is to find the optimal path $\boldsymbol{\pi}^*$ that maximizes $\sum_{t=0}^{\tau} \gamma^t \bcap$, } 
%
%
where $\gamma \in (0,1]$ is a discount factor.


Our risk-aware version of the MESPP problem is an {\em online} problem defined as follows. \bea{Let us associate a ground truth danger level $l_v \in \lset$ for each $v \in V$, and, accordingly, a probability of agent loss for each danger level, $\ploss$. Define for each agent $a \in A$, a fixed nominal danger threshold $\dthres \in \lset$, which is the expected danger level that agent $a$ is apt to endure throughout the mission. We assume the agents are equipped with a danger estimation module they can leverage for estimating danger level distributions $\boldsymbol{\eta}_v^t$, for each vertex $v$ and step $t$}. Such a module provides estimates at $t=0$ simply based on a fixed prior, uniform by default. Once an agent visits a vertex $v$ for the first time, it is allowed to update the distribution of $v$ (for example, by leveraging the method of Section V) and, possibly, that of \bea{neighboring vertices}. We introduce two risk-aware variants of MESPP, based on different additional path constraints, \bea{contingent upon available danger information}.

{\bf Point estimate constraints:} at each step $t$, the current plan of each agent $a$ can not prescribe a visit to a vertex $v$ where the most probable danger level of the vector $\boldsymbol{\eta}_{v}^t$, denoted by $\dangerpt \in \argmax_{\level \in \lset} {\dangerhat}$
%
%
belongs to a danger level strictly larger than $\dthres$.

\bea{{\bf Cumulative probability constraints:} define an agent's required danger confidence, $\dconf \in (0, 1]$ as the cumulative probability of estimated danger up to that agent's threshold $\dthres$. At each step $t$, the current plan of each agent $a$ can only include vertices where the cumulative danger distribution, denoted by $\dangerprob = \sum_{\level = 1}^{\dthres} \eta^t_{l,v} $
%
%
\noindent is equal or higher than $\dconf$. This allows to express more nuanced constraints since we need to consider all danger probabilities up to an agent's danger threshold ${\dthres}$.}

In both cases, the problem objective remains locating the victim as soon as possible, considering that agents might be lost along the mission \bea{according to $\ploss$. Note that while we define our thresholds as static parameters, they can also be time-dependent, allowing for a {dynamic behavior} throughout the mission.}

\subsection{MILP Models}

Our solution is based on a receding-horizon distributed planning approach, where planning is performed by iteratively solving an extension of the MILP models presented in \cite{asfora2020}. We refer the reader to \cite{asfora2020} for the complete set of MILP variables and constraints, as well as implementation details, and focus here solely on modeling danger-related constraints. 

Analogously to the original MILP models, the binary variable $\tost{x}{}_v$ indicates agent $a$ is at vertex $v$ at time $t$ in the computed path. Plans compliant with the point estimated constraints are obtained by enforcing\footnote{We remark that a similar effect of usage could be obtained by removing unsuitable states in the definition of legal searchers' paths, prior to planning (see again~\cite{asfora2020} for details).}
%
\begin{equation}
\tost{x}{}_{v} \dangerpt~ \leq ~\dthres ,~\forall v \in V,~t \in \tset,~a \in A.
\label{eq:point_constraint}
\end{equation} 
%

When dealing with the cumulative probability constraints, we instead impose:
\begin{equation}
\dangerprob \geq~\tost{x}{}_{v}~\dconf,~\forall v \in V,t \in \tset,a \in A.
\label{eq:prob_constraint}
\end{equation} 

\section{Simulations}
\label{sec:experiments}

\subsection{Environment}

We use the School scenario of the DISC dataset \cite{jeon2019disc} with \brev{NFF image distribution (see Sec. \ref{sec:simAssess})} to validate and analyze the performance of our proposed system in its entirety. Since the available dataset measurements represent what the robot would have collected while moving through the simulated environment, we use the left camera poses to infer our layout, which is then abstracted into the graph shown in Fig. \ref{fig:graph}. \brev{We also use it to re-build the environment in Gazebo \cite{gazebo}, an open-source 3D robotics simulator, for more realistic experiments accounting for robot dynamics, asynchronous agents and navigation challenges. } 

\subsection{Danger Estimation}

The general setup is the same for both numerical \brev{and qualitative simulations. The human-reasoned ground truth} for each vertex (Fig. \ref{fig:graph}, left), defines the probability of losing the agent $\ploss = 7.47e-8(\level_v-1)\mathrm{e}^{\level_v} ~\forall l_v \in \lset$, which yields values between 0.009\% and 49.5\%. 


A partial collection of images (5\%) is used for danger distribution estimation. \brev{These correspond to the first set of images an agent would see when entering the vertex area, according to the DISC dataset camera pose. Note from Fig \ref{fig:graph} that the estimated danger distribution, and thus the maximum likelihood level (right), do not always match the human-reasoned ground truth (left).} Processing all images \brev{in a scene} requires minutes even with a powerful GPU \cite{he2017}, \brev{thus} using the \brev{first few acquired} images for estimation aligns well with practical time limitations. 

 \brev{We divide the school space into neighborhoods, based upon common structures such as walls and doorways (see \figc{fig:graph}). We start with an a priori danger distribution for each vertex and} the estimated danger distribution for that vertex is made available when an agent visits it for the first time, mimicking online information gathering. \brev{This information is then spread to vertices in the neighborhood, if they have not been visited yet. For instance, once an agent reaches $v=3$, its danger information ($\boldsymbol{\eta}_{3}^{t}, z_{3}^{t}, H_{3}^{a, t}$) is passed on to the other vertices in the \textit{gym} neighborhood, $v=4, 5, 6$, but not $v=1,2$ since they have already been visited.}

In our simulations, a discrete time-step comprises the following actions: call the planner; get a new plan for the search team; move the agents towards their next goals; retrieve danger data; update danger estimation \brev{on visited and neighboring vertices; compute and apply probability of loss given ground truth; update team status (if agents are lost);} scan for the victim; and finally, output target detection status.  

\subsection{Planner Parameters}

In order to define a challenging initial belief vector, we pick nine random vertices across neighborhoods and assign a uniform probability of victim location among the chosen vertices, assuming such is static when updating the belief. Our team of three agents starts from $v= 1$ (gym entrance) with probability of capture equal to zero, i.e. the victim is not reachable at $t = 0$. For simplicity, we consider a perfect sensing model \jacopo{with detection when both agent and target are in the same location.} However, assuming different victim motions and sensing models is straightforward \cite{asfora2020}.

We use GUROBI \cite{gurobi2019} on 8 threads of a machine equipped with Intel-Core i9-9900 K and 32 GB RAM to solve our path planning problem. We implement a distributed approach, with a mission deadline and planning horizon of 100 and 14 time-steps, respectively. \jacopo{The solving time is consistently under 0.1 secs for the settings studied in this paper.}


\brev{\subsection{Configurations}
\label{sec:config}

We perform ten sets of experiments, with 1000 instances each. We vary: 
 	\subsubsection{planner} without danger constraints (NC), with point estimate (PT) and cumulative probability constraints (PB);
 	\subsubsection{a \textbf{p}riori danger knowledge} available to our agents at $t=0$, either perfect (ground truth) \textbf{k}nowledge (PK) or no knowledge at all, where we assume an \textbf{u}niform distribution for all vertices (PU);
 	\subsubsection{team \makeup} different threshold combinations; (345) for $\boldsymbol{\kappa} = [3, 4, 5]$, $\boldsymbol{\alpha} = [0.6, 0.4, 0.4]$; (335) $\boldsymbol{\kappa} = [3, 3, 5]$, $\boldsymbol{\alpha} = [0.6, 0.6, 0.4]$; and (333) $\boldsymbol{\kappa} = [3, 3, 3]$, $\boldsymbol{\alpha} = [0.6, 0.6, 0.6]$; 
 	\subsubsection{best case baseline} danger-free environment (ND), i.e., without probability of loss.
 
 The configurations are denoted in this order: \{planner -- a priori knowledge -- team makeup\}. Thus, PB-PK-345 denotes cumulative probability planner, perfect a priori knowledge and team threshold makeup of $\boldsymbol{\kappa} = [3, 4, 5]$, $\boldsymbol{\alpha} = [0.6, 0.4, 0.4]$.}

\brev{\subsection{Metrics}}

\subsubsection{Mission outcomes} success, target is found within the deadline; abort, all agents are lost; and cutoff, deadline is reached before target is found.

\subsubsection{Average mission time} discrete time step when mission ends, either due to target detection, mission abortion or cutoff.

\subsubsection{Losses} \brev{percentage of missions where agents are lost due to the dangerous environment. To evaluate the safety potential of our framework, we denote as `Most Valuable Agent' (MVA) the agent we want to protect the most, setting its danger threshold as the lowest among the team. Thus, for the team configuration (345), the MVA is agent $a=1$ with $\kappa^{\rm MVA} = 3$; for (335), there are two MVAs, $a=1,2$; configuration (333) represents a homogeneous team. For a fair comparison, we consider a MVA loss when \textit{at least} one MVA is lost (analogously for N-MVA).}

  \begin{figure*}[h!]
 	\centering
 	\includegraphics[width=0.3\textwidth]{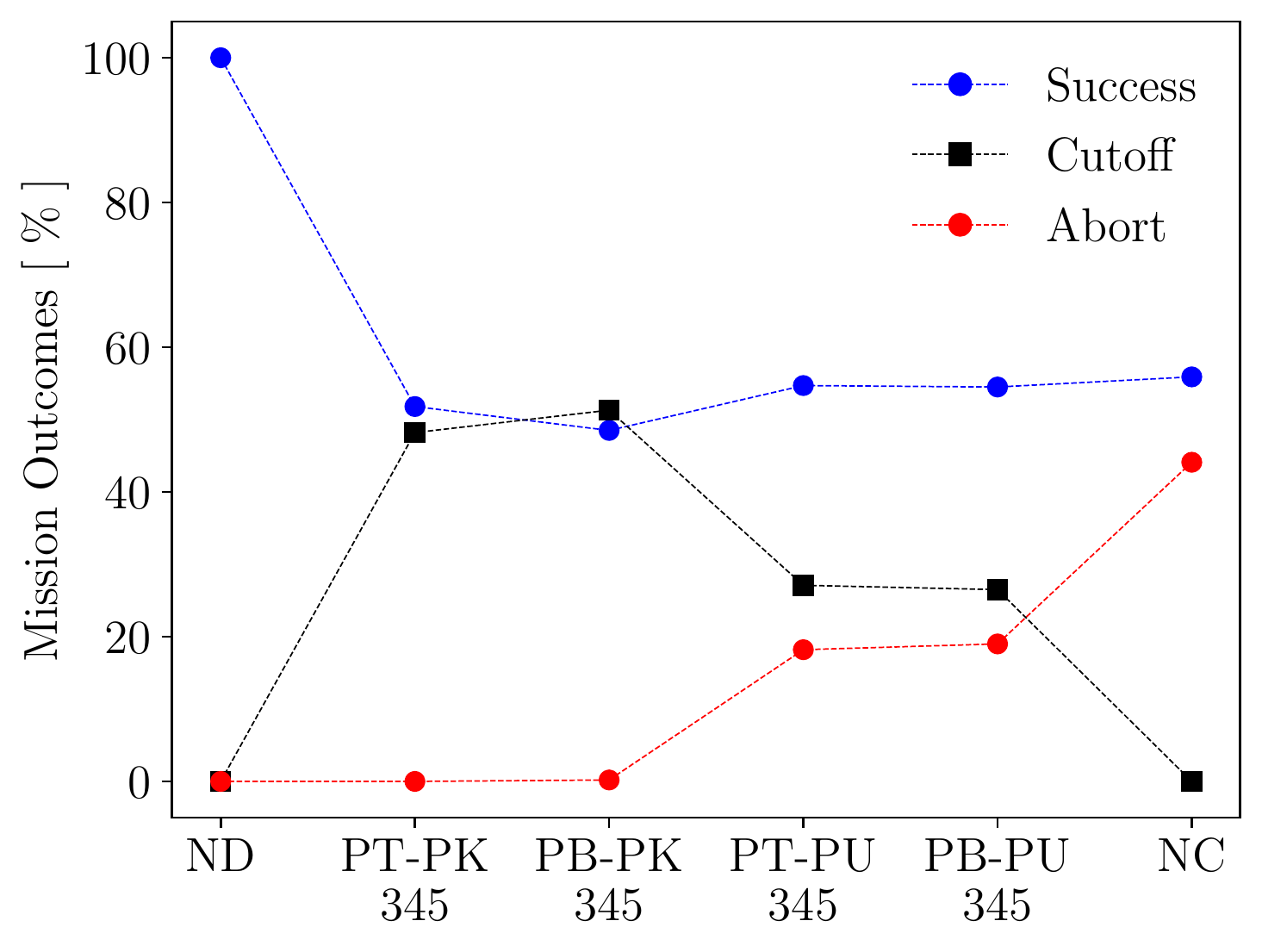}
 	\includegraphics[width=0.3\textwidth]{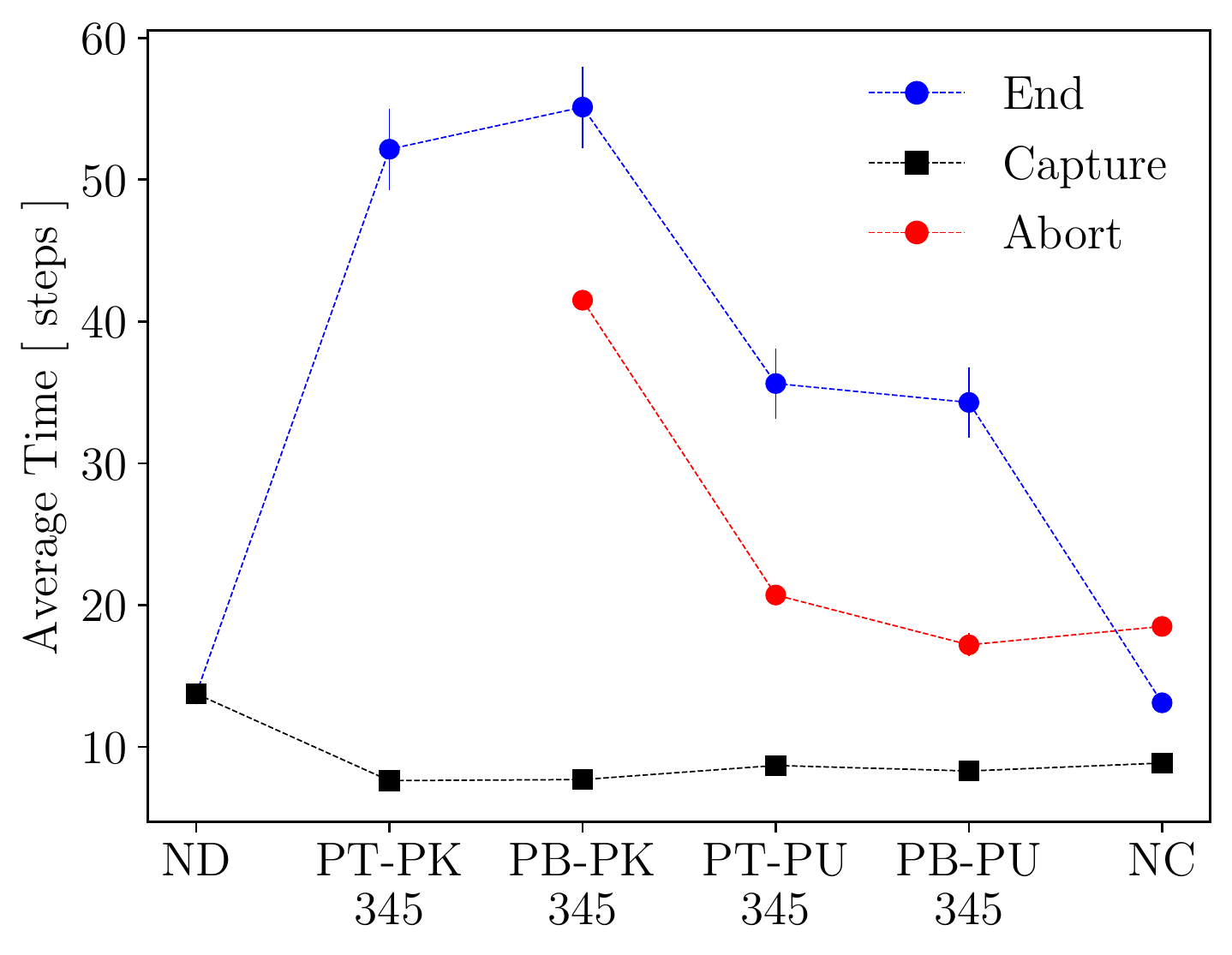}
 	\includegraphics[width=0.3\textwidth]{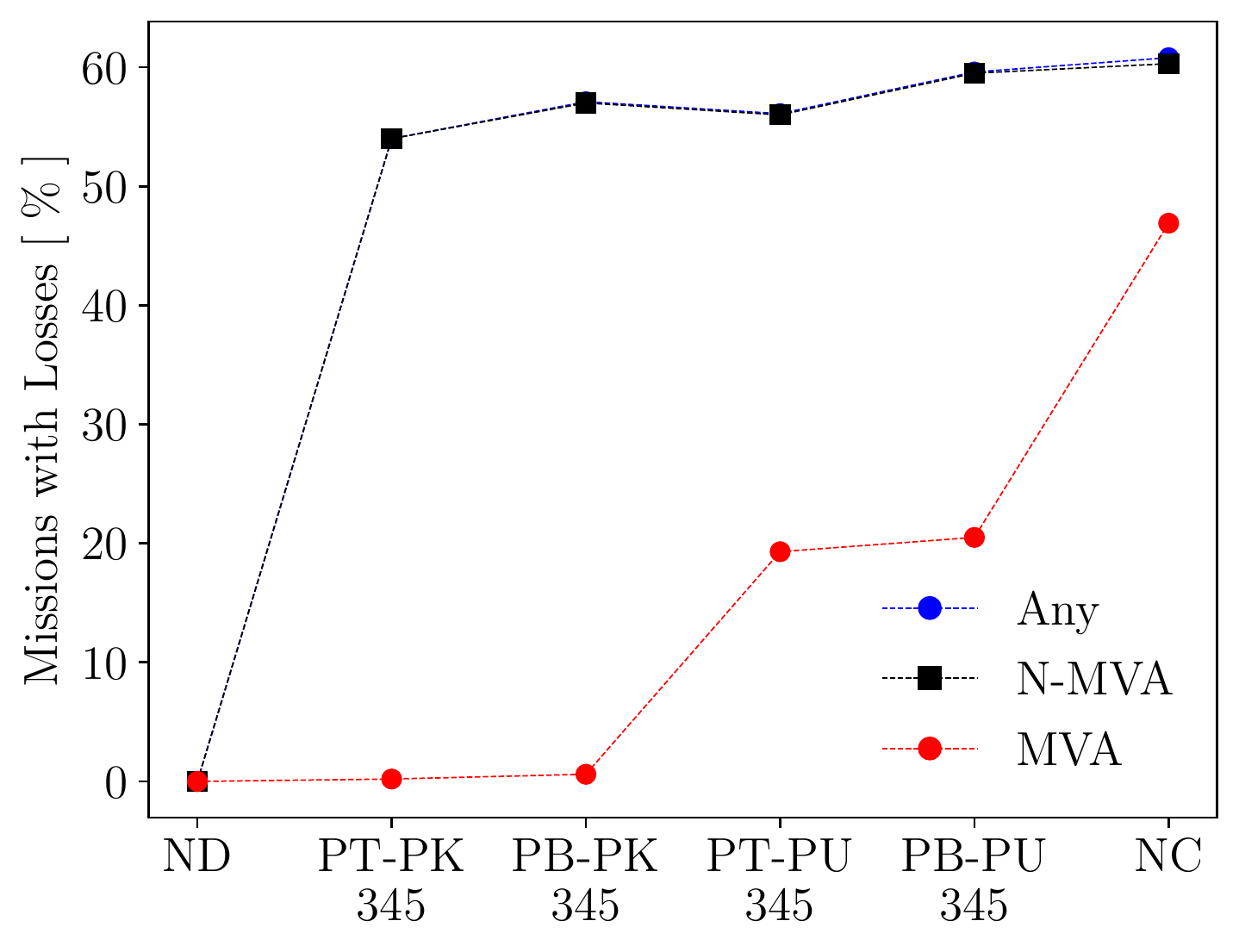}
 	\caption {\small Numerical simulations with varying \textit{planner} and \textit{a priori danger knowledge.}. Planner: point estimate (PT) and cumulative probability (PB) constraints, with $\boldsymbol{\kappa} = [3, 4, 5]$ and $\boldsymbol{\alpha} = [0.6, 0.4, 0.4]$. A priori knowledge: uniform (PU) and perfect (PK), i.e., equal to the human-reasoned ground truth. Additionally, we simulate best (danger-free environment, ND) and worst case (no constraints, NC) scenarios. Left: Mission outcomes.  Middle: Mission times. Right: Losses.}
 	\label{fig:num_sims}
 	\vspace{-0.1in}
 \end{figure*}

 \subsection{Numerical Results}

The numerical simulation results with varying planner and a priori knowledge are shown in Fig. \ref{fig:num_sims}, for an heterogeneous team with danger thresholds $\boldsymbol{\kappa} = [3, 4, 5]$ and $\boldsymbol{\alpha} = [0.6, 0.4, 0.4]$. 

\brev{
The best-case scenario (ND) establishes a baseline of 100\% success in the environment studied, which starts to decrease when probabilities of loss and danger constraints are added. The former has the effect of incapacitating the team so they are unable to proceed with the mission, while the later may slow down the search if some plans are considered unsuitable given the agents thresholds.

With perfect knowledge (PT-PK, PB-PK), the team is able to plan accordingly (\figc{fig:num_sims}, left-red): there are only two missions aborted, and our MVA is rarely lost (\figc{fig:num_sims}, right-red). However, as danger estimation becomes less accurate (PT-PU, PB-PU), the planner cannot be so protective, thus the MVA losses increase (\figc{fig:num_sims}, right-red) as well as aborted missions (\figc{fig:num_sims}, left-red), leading to the worst-case scenario, where danger is present but there are no constraints (NC). Note the cause of failure with perfect knowledge (PK) is due to the deadline being reached, while without danger constraints (NC) the mission fails due to abort, which is corroborated by its much shorter average mission time\footnote{Bars denote 95\% confidence interval; if not shown, values are $\leq 1$ times-step and overlap with marker.} (\figc{fig:num_sims}, middle-blue). The target detection time is roughly the same across configurations  (\figc{fig:num_sims}, middle-black), indicating efficiency is still the main goal. Whenever this goal is not achieved, however, the danger constraints still allow the agents to be safer. Meanwhile, if danger constraints are not used and the mission fails, both agents and target are lost.}

\begin{figure}[ht]
    \centering
    \vspace{0.2cm}
    \includegraphics[width=0.491\columnwidth]{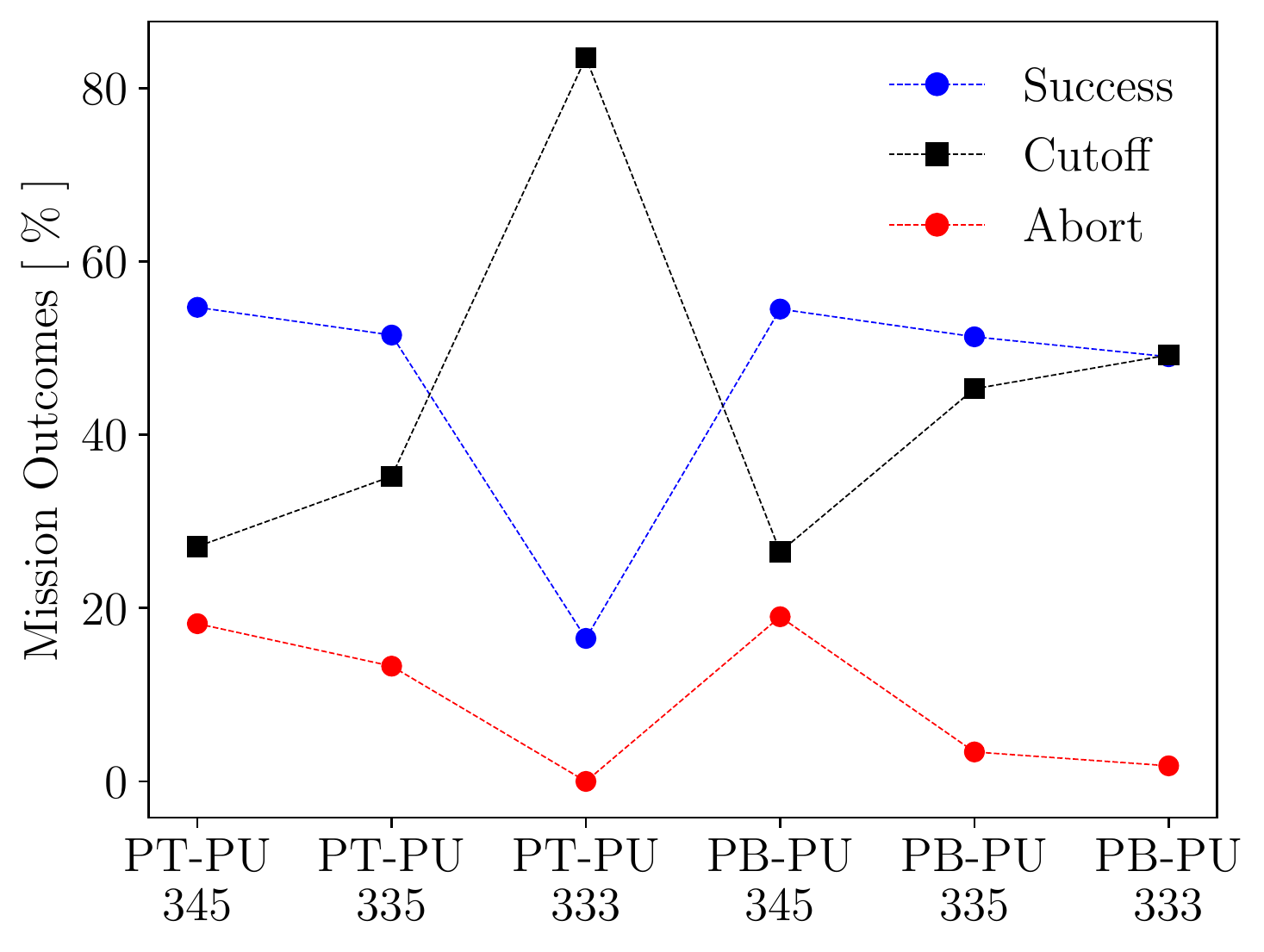}
    \includegraphics[width=0.491\columnwidth]{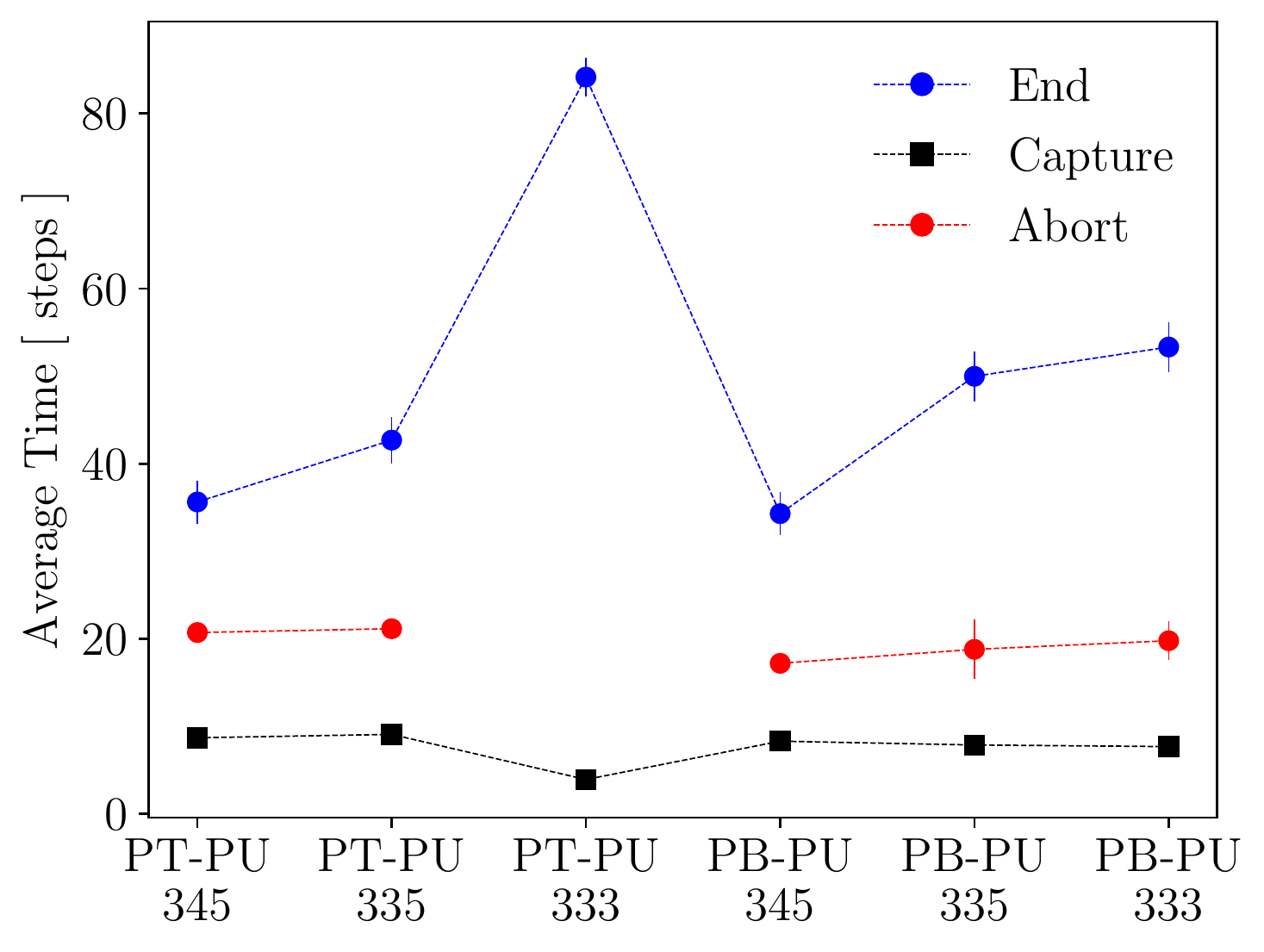}
    \caption{\brev{Numerical  simulations  with  varying team threshold \makeup s. Left: Mission outcomes. Right: Mission times.}} 
    \label{fig:ht_team}
\end{figure}

\brev{Results for different team threshold \makeup s are shown in Fig. \ref{fig:ht_team}. The cumulative probabilistic constraints (PB) have a more stable performance than the point estimate (PT) with different teams, as seen in Fig. \ref{fig:ht_team} (left). For a homogeneous team (PB-PU-333), the mission success rate is only slightly lower than for a heterogeneous one (PB-PU-345). This is likely due to the more nuanced thresholding, e.g. 60\% confidence that a vertex level is between 1-3, rather than considering a single estimate. However, point estimate is simpler to tune and therefore may perform better in a multimodal danger distribution. For both planners, the main cause of failure is cutoff, particularly for PT-PU-333, which presents a significant longer mission time (Fig. \ref{fig:ht_team}, right), and lower success rate. On the other hand, the mission abort rate decreases as the team becomes more homogeneous, illustrating the trade-off between protecting the agents and exploring the environment. }
%

\brev{\subsection{Qualitative Simulations}


We use Gazebo 7 \cite{gazebo} with ROS-Kinetic \cite{ros} to incorporate one Hector quadrotor (agent 1) and two Jackal ground robots\footnote{ROS packages: \url{https://github.com/tu-darmstadt-ros-pkg/hector_quadrotor}; \url{https://clearpathrobotics.com/jackal-small-unmanned-ground-vehicle} } (agents 2, 3), as shown in Fig. \ref{fig:gazebo} (left). 
The simulator's main capabilities of mapping, localization, and autonomous navigation are built off the ROS Navigation Stack. Gazebo simulates the actual robot dynamics, which adds to the possibility of mission failure. The robots can now become inactive either due to danger, or incidents while navigating between vertices, for example getting lost, crashing, or tipping over. If a robot cannot reach its goal vertex before a given time has elapsed, that robot is considered to be inactive, and can no longer participate in the search.}


\brev{We perform 5 instances per configuration, with each discrete time step corresponding in average to 11.72 secs. The team performance follows the trend of the numerical simulations, with similar capture time across configurations. The best-case scenario (ND) presents 100\% success, followed by 80\% for heterogeneous teams (PT/PB-PU-345), 60\% for perfect a priori knowledge (PT/PB-PK-345), and 40\% for the others (PT-PU-335, PB-PU-335/333); particularly for PT-PU-333, none of the missions are successful due to cutoff. There are less agent losses and no abort missions with PK, but also less exploring, which results in a lower success rate than PU in these particular instances. All mission failures when employing constraints are due to cutoff, except for one abort instance in PB-PU-345/335. 

Figure \ref{fig:gazebo_sim} shows the percentage of missions where each agent becomes inactive, partitioned into losses caused by Danger and Navigation. There is an expected increase in agent loss when accounting for navigation, particularly for agents 2 and 3 -- which can be attributed to the fact that these are ground robots, and thus more likely to run into each other or get stuck. Our results suggest that a risk-aware planner can reduce losses on one set of agents, without a significant loss in performance, even when dealing with practical navigation challenges and imperfect scene knowledge.}

\brev{ \begin{figure}[ht]
	\centering
	\includegraphics[width=\mywc, height=\myha]{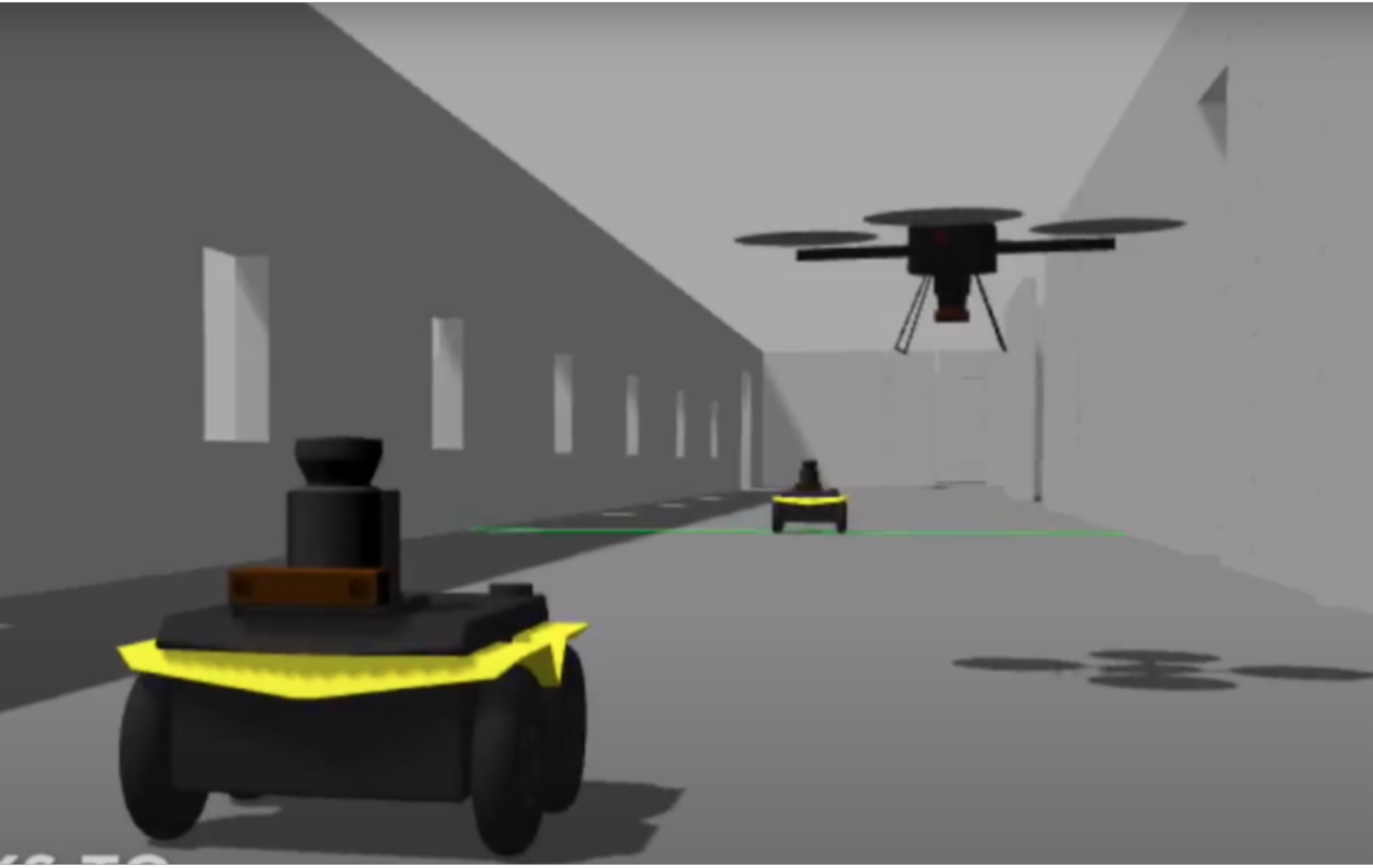}
	\includegraphics[width=\mywc, height=\myha]{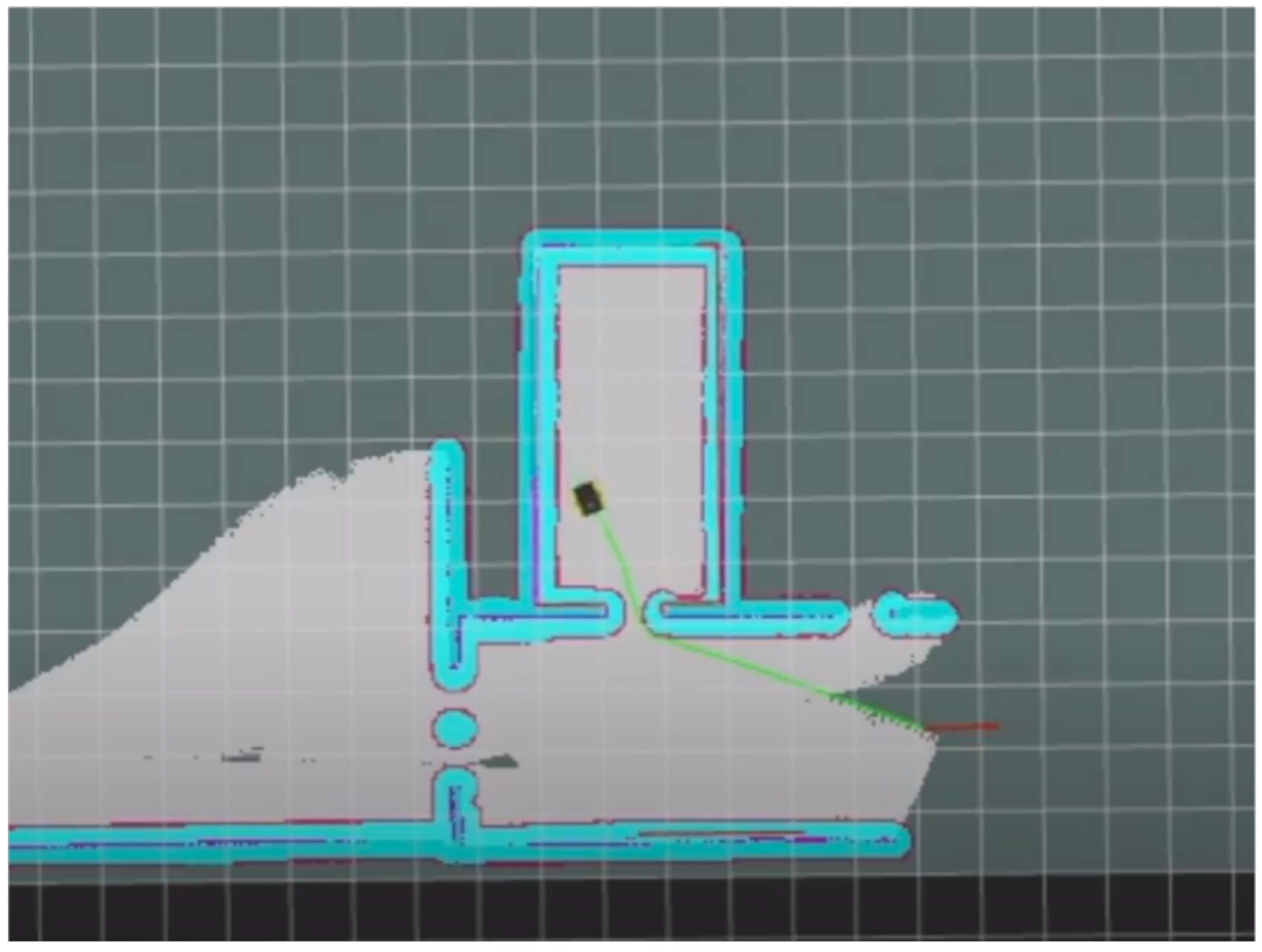}
	\includegraphics[width=\mywc, height=\myha]{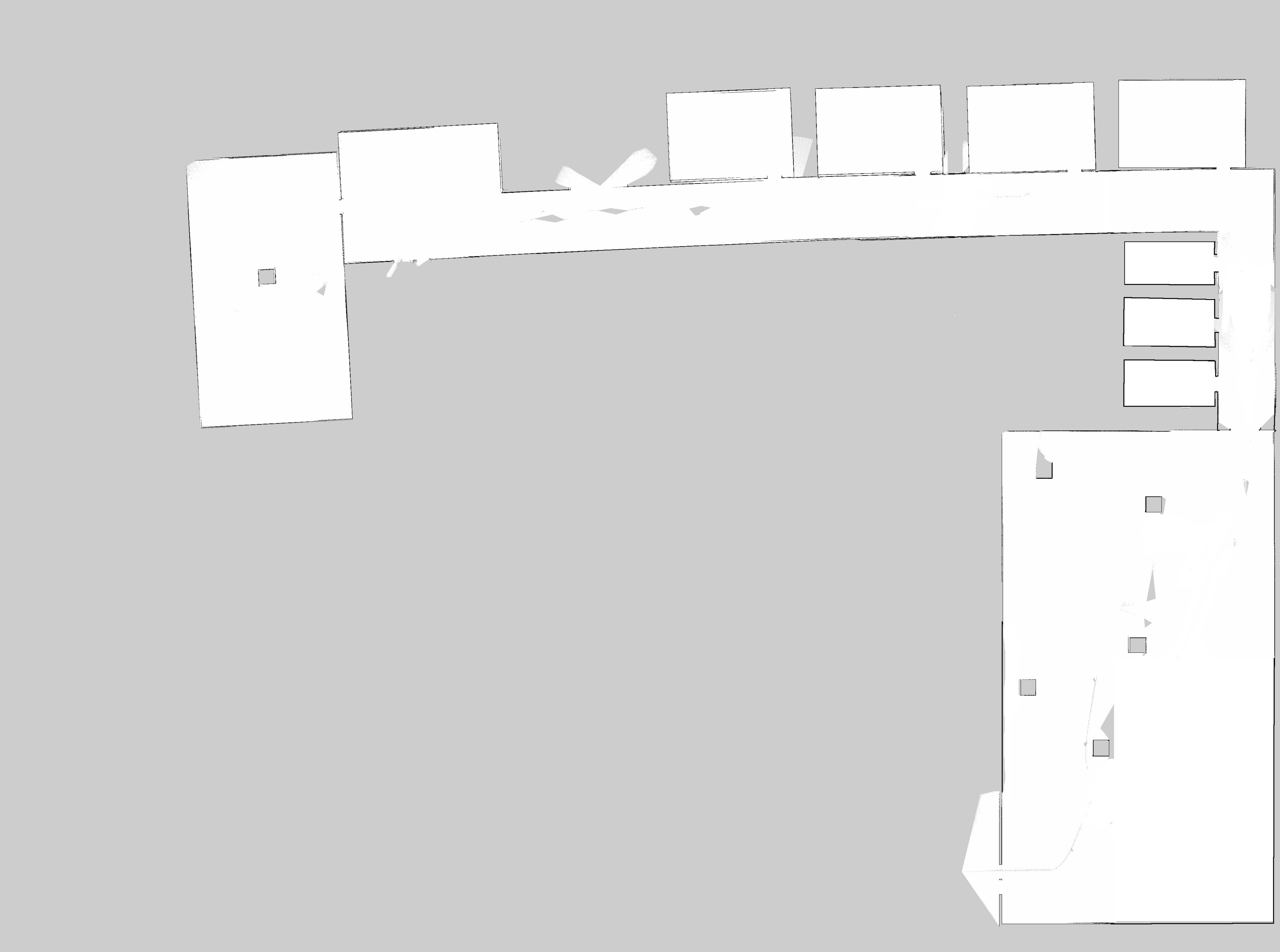}
	\caption{\small \brev{Environment built in ROS/Gazebo based on the poses extracted from DISC dataset. Left: Structure.  Middle: Mapping process. Right: Resultant map.} }
	\label{fig:gazebo}
\end{figure}}
\brev{
\begin{figure}[ht]
    \centering
    \includegraphics[width=1\columnwidth]{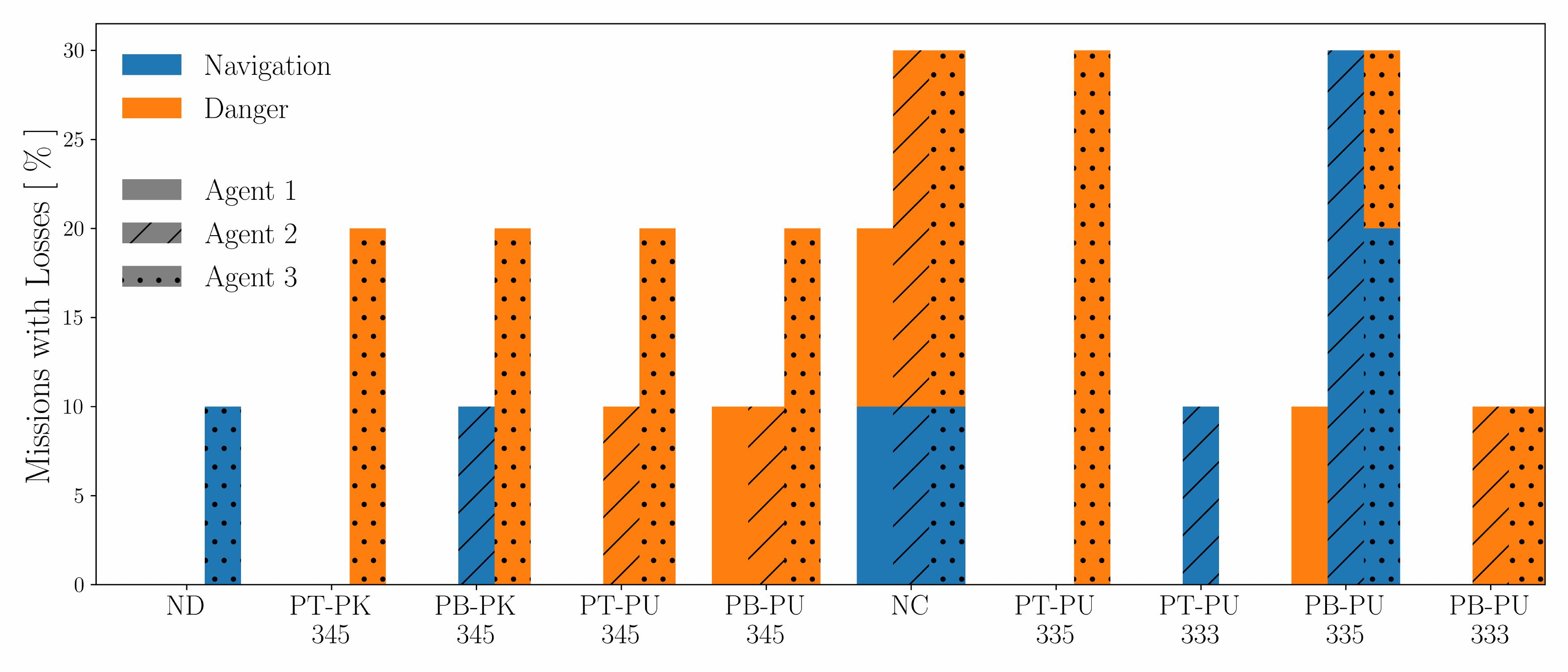}
    \caption{\small \brev{ROS/Gazebo simulations: percentage  of  missions  where  each agent  becomes  inactive, due to dangerous environment or navigation challenges.} }
    \label{fig:gazebo_sim}
\end{figure}}

\newpage
\section{Conclusion}
\label{sec:conclusion}

\jacopo{In this paper, we presented a }danger estimation framework which is adaptable to different environmental settings. Our results shows its potential in enhancing team performance in SaR missions. In the future, we would like to investigate ways to reject redundant collected images, which could potentially reduce the computational overhead during online operations. \vikram{In order to study a more realistic SaR scenario, we would also like to apply a dynamic Bayesian model for estimating danger. We also believe that replacing DISC images with realistic images from movies can be key in bridging the performance gap between our simulations and real-world application.} Furthermore, we would like to tailor our approach to SaR missions with human-robot teams, communicating \bea{succinct and reliable information,} as firefighters do in the real world.







\section*{Acknowledgments}
We thank Asst. Chief Tom Basher (City of Ithaca Fire Dept.) and Chief George Tamborelle (Cayuga Heights Fire Dept.) for their valuable insights on real-world SaR missions.

\bibliographystyle{IEEEtran} 
\bibliography{egbib} 

\end{document}